\documentclass[submission,copyright,creativecommons]{eptcs}
\providecommand{\event}{Doctoral Consortium of International Conference on Logic Programming 2023}
\newenvironment{inlinelist}{\begin{enumerate*}[label=\emph{(\roman{*})}]}
{\end{enumerate*}}
\usepackage[inline]{enumitem}
\usepackage{iftex}

\ifpdf
  \usepackage{underscore}         
  \usepackage[T1]{fontenc}        
\else
  \usepackage{breakurl}           
\fi

\title{Beyond Traditional Neural Networks: Toward adding Reasoning and Learning Capabilities through \\Computational Logic Techniques}
\author{Andrea Rafanelli
\institute{Department of Computer Science\\
University of Pisa,
Pisa, Italy}
\institute{Department of Information Engineering, Computer Science and Mathematics
\\ University of L’Aquila, Italy}
\email{andrea.rafanelli@phd.unipi.it}
}
\def\titlerunning{Beyond Traditional Neural Networks}
\def\authorrunning{Andrea Rafanelli}
\begin{document}
\maketitle

\begin{abstract}
Deep Learning (DL) models have become popular for solving complex problems, but they have limitations such as the need for high-quality training data, lack of transparency, and robustness issues.
Neuro-Symbolic AI has emerged as a promising approach combining the strengths of neural networks and symbolic reasoning. 
Symbolic Knowledge Injection (SKI) techniques are a popular method to incorporate symbolic knowledge into sub-symbolic systems. 
This work proposes solutions to improve the knowledge injection process and integrate elements of ML and logic into multi-agent systems (MAS). 
\end{abstract}

\section{Introduction and Problem Description}

Deep Learning (DL) models have gained popularity for addressing complex problems in various domains.
However, they face challenges. 
Insufficient or biased training data can result in poor model performance and limited generalization. 
DL models heavily rely on statistical models which can restrict their ability to reason about concepts not explicitly represented in the data. 
This limitation hampers their capacity to understand the world and draw accurate conclusions from available information.
To address these issues, Neuro-Symbolic AI has emerged, which seeks to combine the strengths of neural networks and symbolic reasoning to overcome the limitations of purely statistical models. 
There are various techniques for integrating neural and symbolic systems (for a better understanding c.f., e.g., \cite{CalegariCO20,dash2022review,RaedtDMM20}). 
This work focuses mainly on methods that aim to combine these two systems by injecting symbolic knowledge into neural networks, known as Symbolic Knowledge Injection (SKI) techniques.
Additionally, we propose the creation of collaborative networks that combine the strengths of logic-based agents and neural agents within a Neuro-Symbolic architecture, creating a more comprehensive and efficient system.
This approach offers a unique synergy, paving the way for more advanced and intelligent systems.

\section{Background and Existing Literature}
In this section, we will provide a quick overview of the concepts and methods considered in our work.
\paragraph{Symbolic Knowledge Injection}
SKI is a strategy used to improve the performance of sub-symbolic predictors, such as neural networks, by integrating useful symbolic knowledge into them. 
It has several benefits, such as alleviating the problem of insufficient training data, reducing the required time and computational resources for learning, increasing the predictive accuracy of sub-symbolic predictors, and providing human-interpretable frameworks. 
There are several ways the knowledge can be injected into predictors, mainly: 
\begin{inlinelist}
    \item \textit{loss constraining} \cite{Bader2008,DiligentiRG17,XuZFLB18}, i.e., the inclusion of penalty terms in the loss function. Penalty terms are indicative of the limitations imposed by prior knowledge;
    \item \textit{structure constraining} \cite{badreddine22,ManhaeveDKDR21,YuL21}, i.e., the alteration of the structure of the sub-symbolic model to reflect symbolic knowledge expressed as constraints, e.g. by changing the number and size of hidden layers or by setting the connections between neurons;
    \item \textit{knowledge embedding} \cite{BordesUGWY13,ChoiBSSS17,WangWG15}, i.e., incorporating additional domain-specific knowledge by generating numeric data from the symbolic ones. 
\end{inlinelist}

\paragraph{Abductive Logic Programming} 
ALP is a programming paradigm \cite{kakas1992} that combines logic programming with abductive reasoning \cite{peirce} to generate hypotheses (or abducibles) that explain certain observations. 
Abduction is formally described \cite{mooney2000} as a framework where, given a knowledge base $KB$ and observations $O$, one seeks to find a hypothesis $H$, s.t. $KB \cup H \models O.$

ALP is a version of traditional logic programming in which the reasoner generates abductive hypotheses by assuming the truth of abducibles (a set of open predicates) given some integrity constraints.
Integrity constraints are logical formulas of the form :-$Body.$, where \textit{Body} is a conjunction of logical literals.
Intuitively, an integrity constraint specifies a set of conditions that must not hold for any valid solution to the problem.
According to \cite{abductionSemantics}, an abductive framework is known as a triple $< P, A, I_c >$, where $P$ is a collection of clauses, $A$ is a set of predicate symbols, called \textit{abducibles}, and $I_c$ is a set of closed formulae.
Clauses in $P$ are of the form $H \gets L_1, L_2,..,L_k, \enspace k\ge 0$, where $H$ is an atom with a predicate symbol not present in $A$ and $L_i$ is a single literal. 
Therefore, $P$ is a general logic program with the limitation that predicates present in $A$ do not have definitions in $P$. 
Here, the idea is to link to an abductive framework a collection of models--commonly called generalized stable models, or GSMs \cite{GSM}-- that are assumed to reflect the framework itself and may be used to characterise abduction within this framework. Hence, any observation has an abductive explanations if it is consistent with at least one of these GSM models.
Let $<P, A, I_c>$ represent an abductive framework, whereas $O$ represents an observation. 
Then $O$ is explained abductively given a set of hypotheses $\Delta$ iff exists a GSM, $M(\Delta)$, s.t. $M(\Delta) \models O.$
In general, the main aim of abduction, given an observation $O$, is to find a set of \textit{abductive 
explanation} $\Delta \subseteq A$, s.t.:
\begin{inlinelist}
    \item $M \models O.$, and
    \item $M \models I_c.$
\end{inlinelist}

\paragraph{DALI}
DALI ~\cite{CostantiniT02,CostantiniT04} is a programming language developed to construct logical agents. 
It is an extension of the Prolog language\footnote{\url{https://en.wikipedia.org/wiki/Prolog}}, and any Prolog program can be considered a DALI agent.
The agent orientation is provided through events as first-class objects in the language, as well as two types of rules: \textit{reactive rules} and \textit{proactive rules}.
Reactive rules permit agents to interact with and respond to their surroundings.
These rules enable the agents to react to external events. 
\textit{External events} are triggered by the environment of the agent and are denoted by the postfix \textbf{E}. 
An agent may choose to react to an external event by employing a reactive rule containing the event in its head. 
Proactive rules, on the other hand, are used by DALI agents to take initiative and start activities when they deem it suitable. 
Using \textit{internal events}, that are denoted by the postfix \textbf{I}, agents can act independently of their environment and other agents.
DALI agents use \textit{actions} to influence their surroundings, potentially in response to external or internal events.
Actions may have pre-conditions, which are specified in the action rule.
Otherwise, they are just an action atom denoted with the postfix \textbf{A}.
Agents can communicate with and affect their environment through actions. 

\section{Goal of the Research}
This work aims to address challenges in the field of DL models by exploring the integration of neural models and computational logic. 
It will focus on two main tools: SKI and MAS. 
The main goals are: 
\begin{inlinelist}
    \item \textbf{RG 1}: use abduction as a way of integrating targeted knowledge into the injection process.
    \item \textbf{RG 2}: develop metrics to assess the goodness-of-fit of injection mechanisms from several perspectives, both qualitative and efficiency-related.
    \item \textbf{RG3}: formalise an integrated MAS in which one or more agents have purely perceptual tasks and other purely logical tasks. 
\end{inlinelist}

\textbf{RG 1} aims to use abduction to improve the SKI process. 
In supervised learning, the observation comprises input-output pairs encoded as a model $f(\mathcal{X}) \to \mathcal{Y}$. 
When using SKI, we want to modify $f(\cdot)$ to align it with symbolic knowledge.
However, selecting the right rules for each output becomes challenging with increasing rule count, causing possible scalability issues.
One solution is to leverage abduction, which helps identify plausible rules for a given output.
Here, abduction can be seen as a rule $Y :- A.$, where $Y$ is the observation (ground-truth) and $A$ are abducibles generated by the reasoner.
The injection procedure can then be modified to incorporate abducibles, making the predictions of $f(\cdot)$ consistent with the provided abducibles.

\textbf{RG 2} aims to investigate different metrics to gain a deeper understanding of the potential benefits and limitations of SKI from multiple perspectives.
In the field of SKI, the effectiveness of the injection procedure is often measured by comparing the performance of the SKI predictor with its counterpart (non-injected predictor), using metrics such as accuracy, F1 score or MSE.
However, these metrics may not fully capture all aspects of knowledge injection, such as whether such injection resulted in a more sustainable predictor in terms of allocated resources or a more robust predictor, and so on.
Therefore, \textbf{RG 2} seeks to study the injection process through a broader lens, including factors such as the sustainability and robustness of the injection procedure.

Finally, the aim of \textbf{RG 3} is to create a MAS, in which ML and logic-based elements can collaborate and interact with each other.
In practice, given a neuro-symbolic architecture, the main objective is to allocate the different modules of the architecture to different collaborating agents.
The integration of ML and logic-based elements into MAS could potentially lead to more efficient and effective decision-making processes.

\paragraph{Results}

We provided a rough idea of \textbf{RG 1} in a conference position paper \cite{aixia2022}, in which we proposed the incorporation of abductive reasoning through two possible frameworks.
The idea is applied to image segmentation to address issues of data scarcity and low robustness.
In this proposal, we suggest the use of either ABL \cite{DaiX0Z19} or a combination of ABL and knowledge injection. 
The article aims to indicate possible uses of abduction within the realm of Neuro-Symbolic integration.

We presented a preliminary idea of \textbf{RG 2} in \cite{skiqs-woa2022}, where we provide the first - to the best of our knowledge - set of Quality-of-Service (QoS) metrics for SKI, with a focus on quantifying robustness and efficiency advantages owing to injection.
In this paper, we offer an initial formulation for the following metrics: 
\begin{inlinelist}
    \item \textit{memory footprint efficiency}, i.e., the gain in model complexity;
    \item \textit{energy efficiency}, i.e., the gain in total energy required to train and deploy a sub-symbolic model;
    \item \textit{latency efficiency}, i.e., improvements in terms of the time required for inference; 
    \item \textit{data efficiency}, i.e., the improvement in terms of the amount of data required to optimise a sub-symbolic model; 
    \item \textit{robustness}, i.e., the capability of the injection mechanism to adapt to variations of input data and knowledge; 
    \item \textit{comprehensibility}, i.e., the capability of the injection mechanism to produce more intelligible models.
\end{inlinelist}
In addition, this work discusses the potential impact of these metrics within the field of MAS, considering that a probable inefficiency of the sub-symbolic models incorporated in the agents affects their functioning in multiple ways (e.g. energy inefficiency, data inefficiency, computational inefficiency, and so on). 

In a subsequent paper, \cite{skiqosJaamas}, we extended and enhanced \textbf{RG 2} by providing a more rigorous mathematical definition and implementation of the following metrics: memory footprint, energy efficiency, latency efficiency, and data efficiency. 
The software tools necessary for the practical use of these metrics were also provided, and the metrics were evaluated on three different datasets using three different injection models\footnote{The repository containing the experiments can be found at the following link: \url{https://github.com/pikalab-unibo/ski-qos-jaamas-experiments-2022}}.
The results provide valuable insights into the performance of different injection predictors on various datasets, highlighting the importance of adopting specific metrics when evaluating them. 
The paper emphasises that the use of these metrics in the context of MAS could be a crucial tool for comparing different predictors and selecting the most suitable one for a given task.

In \cite{mas-woa2022}, an initial approach to symbolic and sub-symbolic integration within a MAS was proposed.
This paper was later extended in \cite{Rafanelli-mas-2023}. These two works are a first step towards \textbf{RG 3}.
In \cite{mas-woa2022}, we present the potential capabilities of an integrated system consisting of logical agents and a neural network specialized in monitoring flood events for civil protection purposes.
The paper describes a framework composed of a group of intelligent agents performing various tasks and communicating with each other to efficiently generate alerts during flood crisis events. 
In \cite{Rafanelli-mas-2023}, the work is extended, and an initial implementation of the framework is provided. 
The paper presents a preliminary prototype of a MAS that autonomously collects weather warnings, categorises related images using a DL module, filters the results, and alerts human operators only if there is a reasonable certainty that a risk situation is occurring.
To this end, we implemented a system\footnote{The repository containing the experiments can be found at the following link: \url{https://github.com/AAAI-DISIM-UnivAQ/MAS_Py_FLOOD}} using a combination of logical agents and a DL component.
The neural network is trained on eight classes of topographic entities to segment the images.
Once the images have been segmented, a \textit{Logical Image Descriptor} (LID) is used to generate a logical description of the segmented mask predicates. 
This description is then submitted to a logical agent that performs the reasoning.
In our system, the reasoning is performed using a perception-fusion approach, where the MAS agents use their perceptions to reason about the environment and make decisions collectively. 
The paper emphasises the adoption of agents based on computational logic with a logical basis to provide verifiability, explainability, and reliability.

\section{Future Works}

In the future, potential research directions include:
\begin{inlinelist}
    \item for \textbf{RG 1}: explore abductive reasoning technique to enhance knowledge injection efficiency and effectiveness. 
    Evaluate the framework's performance through extensive experiments in various domains;
    \item for \textbf{RG 2}: refine and expand evaluation metrics, including comprehensibility metrics, for better assessment of injected models.
    Validate metrics across different scenarios and datasets;
    \item for \textbf{RG 3}: explore collaborative frameworks and hybrid models combining logical and neural reasoning. 
    Possibly evaluate their performance in real-world applications.
\end{inlinelist}  

Overall, these future works aim to improve knowledge injection, evaluation methodologies, and the integration of logical and neural agents, contributing to more efficient and interpretable machine learning models.   

\section*{Acknowledgements}

This PhD work is conducted under the supervision of professors Stefania Costantini (University of L'Aquila, Italy), Fosca Giannotti (Scuola Normale Superiore, Pisa, Italy), and Andrea Omicini (Alma Mater Studiorum--University of Bologna, Italy).

\bibliographystyle{eptcs}

\end{document}